\documentclass[11pt,a4paper]{article}
\usepackage[hyperref]{acl2020}
\usepackage{times}
\usepackage{latexsym}

\usepackage{microtype}

\usepackage{linguex}
\usepackage{amsmath,amssymb}
\usepackage{multirow,makecell}
\usepackage[shortlabels]{enumitem}
\usepackage[skip=0pt]{subcaption}
\usepackage{booktabs}
\usepackage{comment}
\usepackage[colorinlistoftodos,prependcaption,textsize=tiny]{todonotes}

\aclfinalcopy 
\def\aclpaperid{1665} 

\title{Analysing Lexical Semantic Change\\ with Contextualised Word Representations}

\author{Mario Giulianelli  \qquad Marco Del Tredici \qquad Raquel Fern\'{a}ndez\\
  Institute for Logic, Language and Computation \\
  University of Amsterdam \\
  \texttt{\{m.giulianelli|m.deltredici|raquel.fernandez\}@uva.nl} \\}

\date{}

\begin{document}
\maketitle

\begin{abstract}
This paper presents the first unsupervised approach to lexical semantic change that makes use of contextualised word representations. We propose a novel method that exploits the BERT neural language model to obtain representations of word usages, clusters these representations into usage types, and measures change along time with three proposed metrics. We create a new evaluation dataset and show that the model representations and the detected semantic shifts are positively correlated with human judgements. Our extensive qualitative analysis demonstrates that our method captures a variety of synchronic and diachronic linguistic phenomena. We expect our work to inspire further research in this direction.
\end{abstract}


\section{Introduction}
\label{sec:introduction}

In the fourteenth century the words \textit{boy} and \textit{girl} referred respectively to a male servant and a young person of either sex (Oxford English Dictionary). By the fifteenth century a narrower usage had emerged for \textit{girl}, designating exclusively female individuals, whereas by the sixteenth century \textit{boy} had lost its servile connotation and was more broadly used to refer to any male child, becoming the masculine counterpart of \textit{girl} \citep{bybee2015language}. Word meaning is indeed in constant mutation and, since correct understanding of the meaning of individual words underpins general machine reading comprehension, it has become increasingly relevant for computational linguists to detect and characterise lexical semantic change---e.g., in the form of laws of semantic change \citep{dubossarsky2015category,xu2015laws,hamilton2016diachronic}---with the aid of quantitative and reproducible evaluation procedures \citep{schlechtweg2018durel}. 

Most recent studies have focused on \textit{shift detection}, the task of deciding whether and to what extent the concept evoked by a word has changed between time periods 
\citep[e.g.,][]{gulordava2011,kim2014,kulkarni2015,deltredici2019short,hamilton2016diachronic,bamler2017dynamic,rosenfeld2018deep}.
This line of work relies mainly on distributional semantic models, which 
produce one abstract representation for every word form. However, aggregating all senses of a word into a single representation is particularly problematic for semantic change as word meaning hardly ever shifts directly from one sense to another, but rather typically goes through polysemous stages \citep{hopper1991grammaticization}.
This limitation has motivated recent work on word sense induction across time periods \citep{lau2012,cook2014,mitra2014,frermann2016,rudolph2018,hu2019diachronic}.
Word senses, however, have shortcomings themselves as they are a discretisation of word meaning, which is continuous in nature and  modulated by context to convey ad-hoc interpretations \citep{brugman1988lexicon,kilgarriff1997,paradis2011metonymization}. 

In this work, we propose a usage-based approach to lexical semantic change, where sentential context modulates lexical meaning ``on the fly'' \citep{ludlow2014}.
We present a novel method that (1) exploits a pre-trained neural language model \citep[BERT;][]{devlin2018} to 
obtain contextualised representations for every occurrence of a word of interest, (2) clusters these representations into \emph{usage types}, and (3) measures change along time. More concretely, we make the following contributions: 

\begin{itemize}[leftmargin=11pt,topsep=2pt,itemsep=1pt]
\item We present the first unsupervised approach to lexical semantic change that makes use of state-of-the-art contextualised word representations.
\item We propose several metrics to measure semantic change with this type of representation. Our code is available at \url{https://github.com/glnmario/cwr4lsc}.
\item We create a new evaluation dataset of human similarity judgements on more than 3K word usage pairs across different time periods, available at \url{https://doi.org/10.5281/zenodo.3773250}. 
\item We show that both the model representations and the detected semantic shifts are positively correlated with human intuitions.
\item Through in-depth qualitative analysis, we show that the proposed approach captures synchronic phenomena such as word senses and syntactic functions, literal and metaphorical usage, as well as diachronic linguistic processes related to narrowing and broadening of meaning across time. 
\end{itemize}

\noindent
Overall, our study demonstrates the potential of using contextualised word representations for modelling and analysing lexical semantic change and opens the door to further work in this direction. 



\section{Related Work}
\label{sec:related}

\paragraph{Semantic change modelling}
Lexical semantic change models build on the assumption that meaning change results in the modification of a word's linguistic distribution.
In particular, with the exception of a few methods based on word frequencies and parts of speech \citep{michel2011quantitative, kulkarni2015}, lexical semantic change detection has been addressed following two main approaches: \emph{form-based} and \emph{sense-based} \citep[for an overview, see][]{kutuzov2018survey,tang2018survey}.

In \emph{form-based} approaches independent models are trained on the time intervals of a diachronic corpus and the distance between representations 
of the same word in different intervals is used as a semantic change score \citep{gulordava2011, kulkarni2015}.
Representational coherence between word vectors across different periods can be guaranteed by incremental training procedures \citep{kim2014} as well as by post hoc alignment of semantic spaces \citep{hamilton2016diachronic}.
More recent methods capture diachronic word usage by learning dynamic word embeddings that vary as a function of time \citep{bamler2017dynamic,rosenfeld2018deep,rudolph2018}.
Form-based models depend on a strong simplification: that a single representation is sufficient to model the different usages of a word.

Time-dependent representations are also created in \textit{sense-based} approaches: in this case word meaning is encoded as a distribution over word senses. 
Several Bayesian models of sense change have been proposed \citep{wijaya2011understanding,lau2012,lau2014learning,cook2014}. Among these is the recent SCAN model \citep{frermann2016}, which represents (1) the meaning of a word in a time interval as a multinomial distribution over word senses and (2) word senses as probability distributions over the vocabulary.
The main limitation of sense-based models is that they rely on a bag-of-words representation of context. Furthermore, many of these models keep the number of senses constant across time intervals and require this number to be manually set in advance.

Unsupervised approaches have been proposed that do not rely on a fixed number of senses.
For example, the method for novel sense identification by \citet{mitra2015} represents senses as clusters of short dependency-labelled contexts. Like ours, this method analyses word forms within the grammatical structures they appear. 
However, it requires syntactically parsed diachronic corpora and focuses exclusively on nouns. None of these restrictions limit our proposed approach, which leverages neural contextualised word representations.

\paragraph{Contextualised word representations}
Several approaches to context-sensitive word representations have been proposed in the past. \citet{schutze1998automatic} introduced a clustering-based disambiguation algorithm for word usage vectors, \citet{erk2008structured} proposed creating multiple vectors for the same word and \citet{erk2010exemplar} proposed to directly learn usage-specific representations based on the set of exemplary contexts within which the target word occurs. 

Recently, neural contextualised word representations have gained widespread use in NLP, thanks to deep learning models which learn usage-dependent representations while optimising tasks such as machine translation \citep[CoVe;][]{mccann2017learned} and language modelling (\citealp{dai2015semi}, ULMFiT; \citealp{howard2018universal}, ELMo; \citealp{peters2018}, GPT; \citealp{radford2018,radford2019language}, BERT; \citealp{devlin2018}).
State-of-the-art language models typically use stacked attention layers \citep{vaswani2017attention}, they are  pre-trained on a very large amount of textual data, and they can be fine-tuned for specific downstream tasks \citep{howard2018universal,radford2019language,devlin2018}. 

Contextualised representations have been shown to encode lexical meaning dynamically, reaching high accuracy on, e.g., the binary usage similarity judgements of the WiC evaluation set \citep{wic2018}, performing on a par with state-of-the-art word sense disambiguation models \citep{wiedemann2019does}, and proving useful for the supervised derivation of time-specific sense representation \citep{hu2019diachronic}.
In this work, we investigate the potential of contextualised word representations to detect and analyse lexical semantic change, without any lexicographic supervision.


\section{Method: A Usage-based Approach to Lexical Semantic Change}
\label{sec:method}
We introduce a usage-based approach to lexical semantic change analysis which relies on contextualised representations of unique word occurrences (\textit{usage representations}). First, given a diachronic corpus and a list of words of interest, we use the BERT language model \citep{devlin2018} to compute usage representations for each occurrence of these words.
Then, we cluster all the usage representations collected for a given word into an automatically determined number of partitions (\textit{usage types}) and organise them along the temporal axis. Finally, we propose three metrics to quantify the degree of change undergone by a word.

\subsection{Language Model}
\label{sec:method:lm}
We produce usage representations using the BERT language model \citep{devlin2018}, a multi-layer bidirectional Transformer encoder trained on masked token prediction and next sentence prediction, on the BooksCorpus (800M words) \citep{zhu2015books} and on English text passages extracted from Wikipedia (2,500M words). There are two versions of BERT. For space and time efficiency, we use the smaller \emph{base-uncased} version, with 12 layers, 768 hidden dimensions, and 110M parameters.\footnote{We rely on \textit{Hugging Face}'s implementation of BERT (available at \url{https://github.com/huggingface/transformers}).}

\subsection{Usage Representations}
\label{sec:method:usages}
Given a word of interest $w$ and a context of occurrence $s = (v_1, ..., v_i, ..., v_n)$ with $w = v_i$, we extract the activations of all of BERT's hidden layers for sentence position $i$ and sum them dimension-wise. We use addition because neither concatenation nor selecting a subset of the layers produced notable differences in the relative geometric distance between word representations. 

The set of $N$ usage representations for $w$ in a given corpus can be expressed as the usage matrix $\textbf{U}_w = \left( \textbf{w}_1, \ldots, \textbf{w}_N \right)$. For each usage representation in the usage matrix $\textbf{U}_w$, we store the context of occurrence (a 128-token window around the target word) as well as a temporal label $\textbf{t}_w$ indicating the time interval of the usage. 

\begin{figure}
\centering
\begin{minipage}{.25\textwidth}
\hspace*{-10pt} 
  \includegraphics[width=\linewidth+10pt]{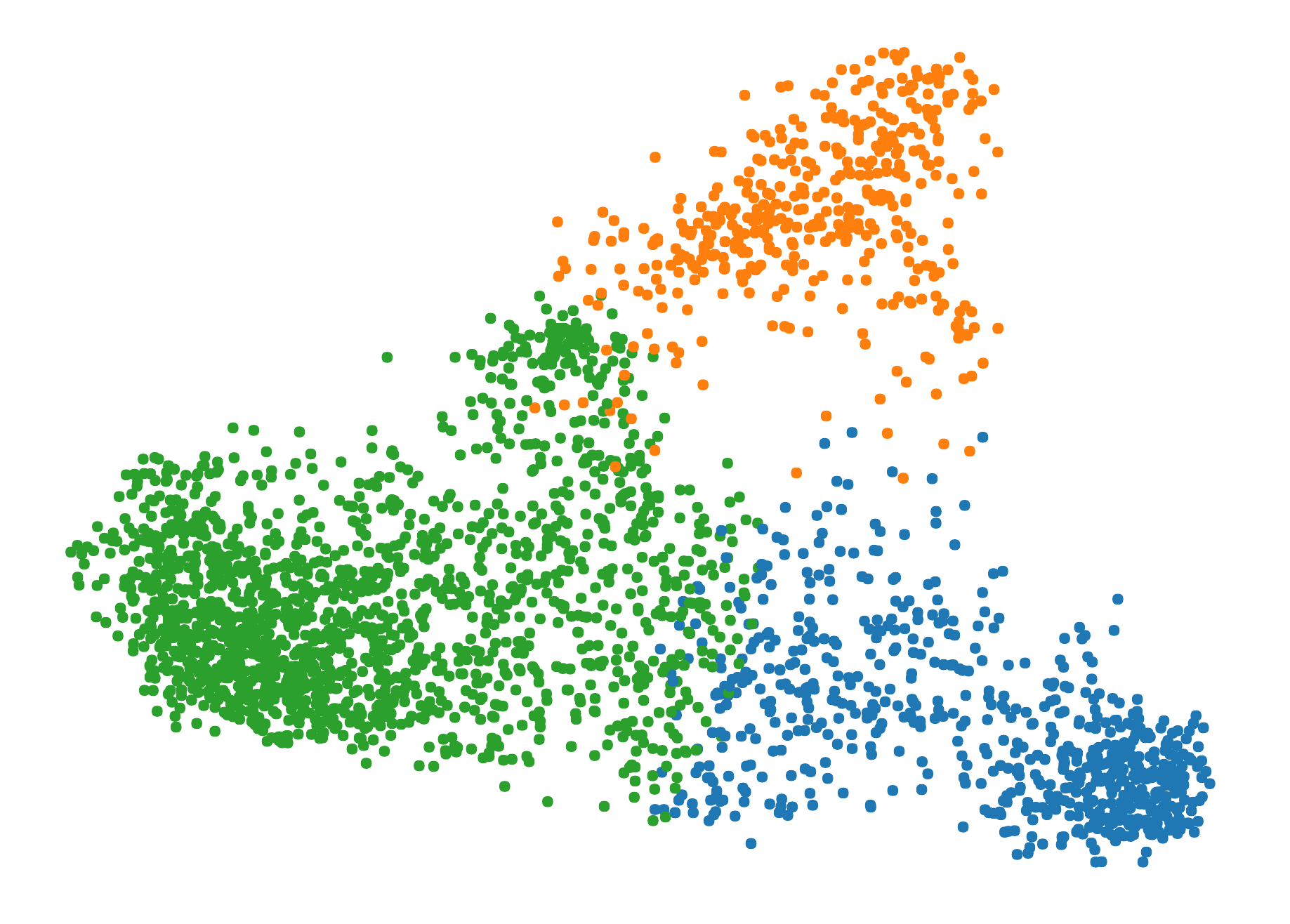}
  \subcaption{PCA visualisation of the\\usage representations.} \label{fig:atom-pca}
\end{minipage}%
\begin{minipage}{.25\textwidth}
\hspace*{-10pt} 
  \includegraphics[width=\linewidth+10pt]{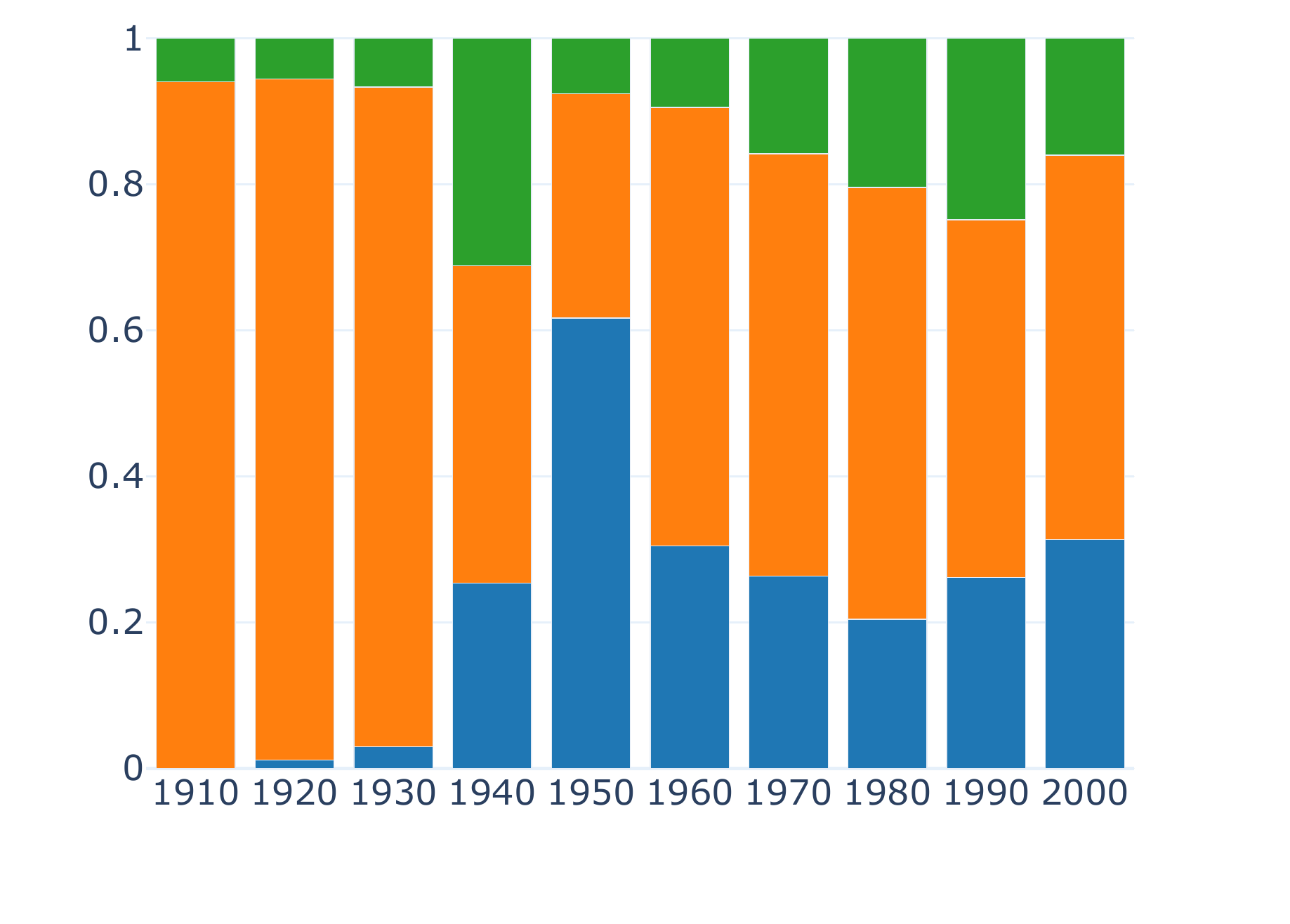}.  
  \subcaption{Probability-based usage\\type distributions along time.} \label{fig:atom-prob}
\end{minipage}
\caption{Usage representations and usage type distributions generated with occurrences of the word \textit{atom} in COHA \protect\citep{davies2012expanding}. Colours encode usage types.} \label{fig:atom}
\end{figure}

\subsection{Usage Types} 
\label{sec:method:clustering}
Once we have obtained a word-specific matrix of usage vectors $\textbf{U}_w$, we standardise it 
and cluster its entries using $K$-Means.\footnote{Other clustering methods are also possible. For this first study, we choose the widely used $K$-Means (\textit{scikit-learn}).} This step partitions usage representations into clusters of similar usages of the same word, or \emph{usage types} (see Figure~\ref{fig:atom-pca}), and thus it is directly related to automatic word sense discrimination \citep[][among others]{schutze1998automatic,pantel2002,manandhar2010semeval, navigli2013semeval}.

For each word independently, we automatically select the number of clusters $K$ that maximises the silhouette score~\citep{rousseeuw1987silhouettes}, a metric of cluster quality which favours intra-cluster coherence and penalises inter-cluster similarity, without the need for gold labels. 
For each value of $K$, we execute 10 iterations of Expectation Maximization to alleviate the influence of different initialisation values \citep{arthur2007k}.
The final clustering for a given $K$ is the one that yields the minimal \textit{distortion} value across the 10 runs, i.e., the minimal sum of squared distances of each data point from its closest centroid. We experiment with $K \in [2, 10]$.
We choose the range $[2, 10]$ heuristically: we forgo $K=1$ as $K$-Means and the silhouette score are ill-defined for this case, while keeping the number of possible clusters manageable computationally.
This excludes the possibility that a word has a single usage type. Alternatively, we could use a measure of intra-cluster dispersion for $K=1$, and consider a word monosemous if its dispersion value is below a threshold $d$ (if the dispersion is higher than $d$, we would discard  $K=1$ and use the silhouette score to find the best $K \geq 2$). There also exist clustering methods that select the optimal $K$ automatically, e.g. DBSCAN or Affinity Propagation \citep{martinc2020}. They nevertheless require method-specific parameter choices which indirectly determine the number of clusters.

By counting the number of occurrences of each usage type \textit{k} in a given time interval \textit{t} (we refer to this count as ${\it freq}(k, t)$), we obtain frequency distributions $\textbf{f}_w^t$ for each interval under scrutiny:
\begin{align}
    \textbf{f}_w^t \in \mathbb{N}^{K_w}: \textbf{f}_w^t[k] = {\it freq}(k, t)\ \ \ k \in [1, K_w]
    \label{eq:distribution}
\end{align}
When normalised, frequency distributions can be interpreted as probability distributions over usage types $\textbf{u}_w^t : \textbf{u}_w^t[k] = \frac{1}{N_t} \textbf{f}_w^t[k]$. Figure~\ref{fig:atom-prob} illustrates the result of this process.

\subsection{Quantifying Semantic Change}
\label{sec:method:metrics}
We propose three metrics for the automatic quantification of lexical semantic change using contextualised word representations.
The first two (\emph{entropy difference} and \emph{Jensen-Shannon divergence}) are known metrics for comparing probability distributions. In our approach, we apply them to measure variations in the relative prominence of coexisting usage types. We conjecture that these kinds of metric can help detect semantic change processes that, e.g., lead to broadening or narrowing (i.e., to increase or decrease, respectively, in the number or relative distribution of usage types). 

The third metric (\emph{average pairwise distance}) only requires a usage matrix $\textbf{U}_w$ and the temporal labels $\textbf{t}_w$ (Section~\ref{sec:method:usages}). Since it does not rely on usage type distributions, it is not sensitive to possible errors stemming from the clustering process.

\paragraph{Entropy difference (ED)}
We propose measuring the uncertainty (e.g., due to polysemy) in the interpretation of a word $w$ in interval $t$ using the
normalised entropy of its usage distribution~$\textbf{u}_w^t$:
\begin{align}
\eta(\textbf{u}_w^t) = \log_{K_w} \left(\ \prod_{k=1}^{K_w} \textbf{u}_w^t[k]^{- \textbf{u}_w^t[k]}\right)
\end{align}
To quantify how uncertainty over possible interpretations varies across time intervals, we compute the difference in entropy between the two usage type distributions in these intervals: $\operatorname{ED}(\textbf{u}_w^t, \textbf{u}_w^{t'}) = \eta(\textbf{u}_w^{t'}) - \eta(\textbf{u}_w^{t})$. 
We expect high ED values to signal the broadening of a word's interpretation and negative values to indicate narrowing. 

\paragraph{Jensen-Shannon divergence (JSD)}
The second metric takes into account not only variations in the size of usage type clusters but also \textit{which clusters} have grown or shrunk. It is the Jensen-Shannon divergence \citep{lin1991jsd} between usage type distributions:
\begin{equation}
\begin{aligned}
\operatorname{JSD}(\textbf{u}_w^t, \textbf{u}_w^{t'}) &= \operatorname{H}\left(\frac{1}{2} \left( \textbf{u}_w^t + \textbf{u}_w^{t'}\right) \right) \\
&- \frac{1}{2} \left(\operatorname{H}\left(\textbf{u}_w^t\right) - \operatorname{H}\left(\textbf{u}_w^{t'}\right)\right)
\end{aligned}
\end{equation}
where $\operatorname{H}$ is the Boltzmann-Gibbs-Shannon entropy.
Very dissimilar usage distributions yield high JSD whereas low JSD values indicate that the proportions of usage types barely change across periods.

\paragraph{Average pairwise distance (APD)}
While the previous two metrics rely on usage type distributions, it is also possible to quantify change bypassing 
the clustering step into usage types, e.g. by calculating the average pairwise distance between usage representations in different periods $t$ and $t'$: 
\begin{align}
    \operatorname{APD}(\textbf{U}_w^{t}, \textbf{U}_w^{t'}) = \frac{1}{N^t \cdot N^{t'}} \sum_{\textbf{x}_i \in \textbf{U}_w^{t},\ \textbf{x}_j \in \textbf{U}_w^{t'}} d(\textbf{x}_i, \textbf{x}_j)
\end{align}
where $\textbf{U}_w^{t}$ is a usage matrix constructed with occurrences of $w$ only in interval $t$.
We experiment with cosine, Euclidean, and Canberra distance.

\paragraph{Generalisation to multiple time intervals}
\label{sec:method:intervals}
The presented metrics quantify semantic change across pairs of temporal intervals ($t,t'$). When more than two intervals are available, we measure change across all contiguous intervals ($m(\textbf{U}_w^{t}, \textbf{U}_w^{t+1})$, where $m$ is one of the metrics), and collect these values into vectors. 
We then transform each vector into a scalar change score by computing the vector's mean and maximum values.\footnote{The Jensen-Shannon divergence can also be measured with respect to $T > 2$ probability distributions \citep{re2014generalization}: $\operatorname{JSD}\left(\textbf{u}_w^1, \ldots, \textbf{u}_w^{T}\right)=\operatorname{H}\left( \frac{1}{T} \sum_{i=1}^{T} \textbf{u}_w^i \right)- \frac{1}{T} \sum_{i=1}^{T} \operatorname{H}\left(\textbf{u}_w^i\right)$. However, this definition of the JSD is insensitive to the order of the temporal intervals and yields lower correlation with human semantic change ratings (cfr. Section \ref{sec:eval:change}) than the pairwise metrics.} Whereas the \textit{mean} is indicative of semantic change across the entire period under consideration, the \textit{max} pinpoints the pair of successive intervals where the strongest shift has occurred.


\section{Data}
\label{sec:data}
We examine word usages in a large diachronic corpus of English, the Corpus of Historical American English \citep[COHA,][]{davies2012expanding}, which covers two centuries (1810--2009) of language use and includes a variety of genres, from fiction to newspapers and popular magazines, among others.
In this study, we focus on texts written between 1910 and 2009, for which a minimum of 21M words per decade is available, and discard previous decades, where data are less balanced per decade.

We use the 100 words annotated with semantic shift scores by \citet{gulordava2011} as our target words. These scores are human judgements collected by asking five annotators to quantify the degree of semantic change undertaken by each word (shown out of context) from the 1960's to the 1990's. We exclude \textit{extracellular} as in COHA this word only appears in three decades; all other words appear in at least 8 decades, with a minimum and maximum frequency of 191 and 108,796, respectively. We refer to the resulting set of 99 words and corresponding shift scores as the `GEMS dataset' or the `GEMS words', as appropriate. 

We collect a contextualised representation for each occurrence of these words in the second century of COHA,
using BERT as described in Section~\ref{sec:method:usages}. This results in a large set of usage representations, $\sim$1.3M in total, which we cluster into usage types using $K$-Means and silhouette coefficients (Section~\ref{sec:method:clustering}). We use these usage representations and usage types in the evaluation and the analyses offered in the remaining of the paper.


\section{Correlation with Human Judgements}
\label{sec:evaluation}
Before using our proposed method to analyse language change, we assess how its key components compare with human judgements. We test whether the clustering into usage types reflects human similarity judgements (Section~\ref{sec:eval:similarities}) and to what extent the degree of change computed with our metrics correlates with shift scores provided by humans (Section~\ref{sec:eval:change}).

\subsection{Evaluation of Usage Types} 
\label{sec:eval:similarities}
The clustering of contextualised representations into usage types is one of the main steps in our method (see Section~\ref{sec:method:clustering}). It relies on the similarity values between pairs of usage representations created by the language model. To quantitatively evaluate the quality of these similarity values (and thus, by extension, the quality of usage representations and usage types), we compare them to similarity judgements by human raters. 

\paragraph{New dataset of similarity judgements} 
We create a new evaluation dataset, following the annotation approach of \citet{erk2009investigations, erk2013measuring} for rating pairs of usages of the same word. Since we need to collect human judgements for pairs of usages, annotating the entire GEMS dataset would be extremely costly and time consuming. 
Therefore, to limit the scope of the annotation, we select a subset of words. For each shift score value $s$ in the GEMS dataset, we sample a word uniformly at random from the words annotated with $s$. This results in 16 words. 
To ensure that our selection of usages is sufficiently varied, for each of these words, we sample five usages from each of their usage types (the number of usage types is word-specific) along different time intervals, one usage per 20-year period over the century. All possible pairwise combinations are generated for each target word, resulting in a total of 3,285 usage pairs.

We use the crowdsourcing platform Figure Eight\footnote{\url{https://www.figure-eight.com}, recently acquired by Appen (\url{https://appen.com}).} to collect five similarity judgements for each of these usage pairs. Annotators are shown pairs of usages of the same word: each usage shows the target word in its sentence, together with the previous and the following sentences (67 tokens on average). 
Annotators are asked to assign a similarity score on a 4-point scale, ranging from \textit{unrelated} to \textit{identical}, as defined by \citet{brown-2008-choosing} and used e.g., by \citet{schlechtweg2018durel}.\footnote{The full instructions with examples given to the annotators are available in Appendix \ref{appendix-newdataset}.} 
A total of 380 annotators participated in the task. The inter-rater agreement, measured as the average pairwise Spearman's correlation between common annotation subsets, is 0.59. This is in line with previous approaches such as \citet{schlechtweg2018durel}, who report agreement scores between 0.57 and 0.68.

\paragraph{Results}
To obtain a single human similarity judgement per usage pair, we average the scores given by five annotators.  
We encode all averaged human similarity judgements for a given word in a square matrix.
We then compute similarity scores over pairs of usage vectors output by BERT\footnote{For this evaluation, BERT is given the same variable-size context as the human annotators. Vector similarity values are computed as the inverse of Euclidean distance, because $K$-means relies on this metric for cluster assignments.} to obtain analogous matrices per word and measure Spearman’s rank correlation between the human- and the machine-generated matrices using the Mantel test \citep{mantel1967}. 

We observe a significant ($p < 0.05$) positive correlation for 10 out of 16 words,
with $\rho$ coefficients ranging from 0.13 to 0.45.\footnote{Scores per target word are given in Appendix \ref{appendix-correlation}.} This is an encouraging result, which indicates that BERT's word representations and similarity scores (as well as our clustering methods which build on them) correlate, to a substantial extent, with human similarity judgements. We take this to provide a promising empirical basis for our approach.

\subsection{Evaluation of Semantic Change Scores}
\label{sec:eval:change}
We now quantitatively assess the semantic change scores yielded by the metrics described in Section~\ref{sec:method:metrics} when applied to BERT usage representations and the usage types created with our approach. 
We do so by comparing them to the human shift scores in the GEMS dataset. For consistency with this dataset, which quantifies change from the 1960's to the 1990's as explained in Section~\ref{sec:data}, we only consider these four decades when calculating our scores. Using each of the metrics on representations from these time intervals, we assign a semantic change score to all the GEMS words. We then compute Spearman's rank correlation between the automatically generated change scores and the gold standard shift values.

\paragraph{Results}
Table \ref{table:correlation} shows the Spearman's correlation coefficients obtained using our metrics, together with a frequency baseline (the difference between the normalised frequency of a word in the 1960's and in the 1990's). The three proposed metrics yield significant positive correlations. This is again a very encouraging result regarding the potential of contextualised word representations for capturing lexical semantic change.

As a reference, we report the correlation coefficients with respect to GEMS shift scores documented by the authors of two alternative approaches: the count-based model by \citet{gulordava2011} themselves (trained on two time slices from the Google Books corpus with texts from the 1960's and the 1990's) and the sense-based SCAN model by \citet{frermann2016} (trained on the DATE corpus with texts from the 1960's through the 1990's).\footnote{\citet{gulordava2011} report Pearson correlation. However, to allow for direct comparison, \citet{frermann2016} computed Spearman correlation for that work (see their footnote 7), which is the value we report.}  

For all our metrics, the \textit{max} across the four time intervals---i.e., identifying the pair of successive intervals where the strongest shift has occurred (cfr.\ end of Section \ref{sec:method:intervals})---is the best performing aggregation strategy.
Table~\ref{table:correlation} only shows values obtained with \textit{max} and Euclidean distance for APD, as they are the best-performing options.

It is interesting to observe that APD can prove as informative as JSD and ED, although it does not depend on the clustering of word occurrences into usage types. Yet, computing usage types offers a powerful tool for analysing lexical change, as we will see in the next section.

\begin{table}[t]\small
\centering
\resizebox{\columnwidth}{!}{%
\begin{tabular}{lc}\toprule
Frequency difference                 & 0.068\\
Entropy difference (\textit{max})             & 0.278    \\
Jensen-Shannon divergence (\textit{max})         & 0.276    \\
Average pairwise distance (\emph{Euclidean}, \textit{max})       & 0.285   \\ \midrule
\citet{gulordava2011}  & 0.386     \\
\citet{frermann2016}    &     0.377   \\ \bottomrule
\end{tabular}
} 
\caption{Spearman's $\rho$ correlation coefficients between the gold standard scores in the GEMS dataset and the change scores assigned by our three metrics and a relative frequency baseline. For reference, correlation coefficients reported by previous works using different approaches are also given. All correlations are significant ($p<0.05$) except for the frequency difference baseline.} 
\label{table:correlation}
\end{table}


\section{Analysis}
\label{sec:analysis}
In this section, we provide an in-depth qualitative analysis of the linguistic properties that define usage types and the kinds of lexical semantic change we observe. More quantitative methods (such as taking the top $n$ words with highest JSD, APD and ED and checking, e.g., how many cases of broadening each metric captures) are difficult to operationalise \cite{tang2016semantic} because there exist no well-established formal notions of semantic change types in the linguistic literature. To carry out this analysis, for each GEMS word, we identify the most representative usages in a given usage type cluster by selecting the five closest vectors to the cluster centroid, and take the five corresponding sentences as usage examples.

\subsection{What do Usage Types Capture?}
\label{sec:analysis:synchronic}
We first leave the temporal variable aside and present a synchronic analysis of usage types. Our goal is to assess the interpretability and internal coherence of the obtained usage clusters.

We observe that usage types can discriminate between underlying senses of polysemous (and homonymous) words, between literal and figurative usages, and between usages that fulfil different syntactic roles; plus they can single out phrasal collocations as well as named entities. 

\begin{figure*}[t]
 \centering
 \begin{minipage}{.49\textwidth}
   \centering
   \includegraphics[width=0.95\linewidth]{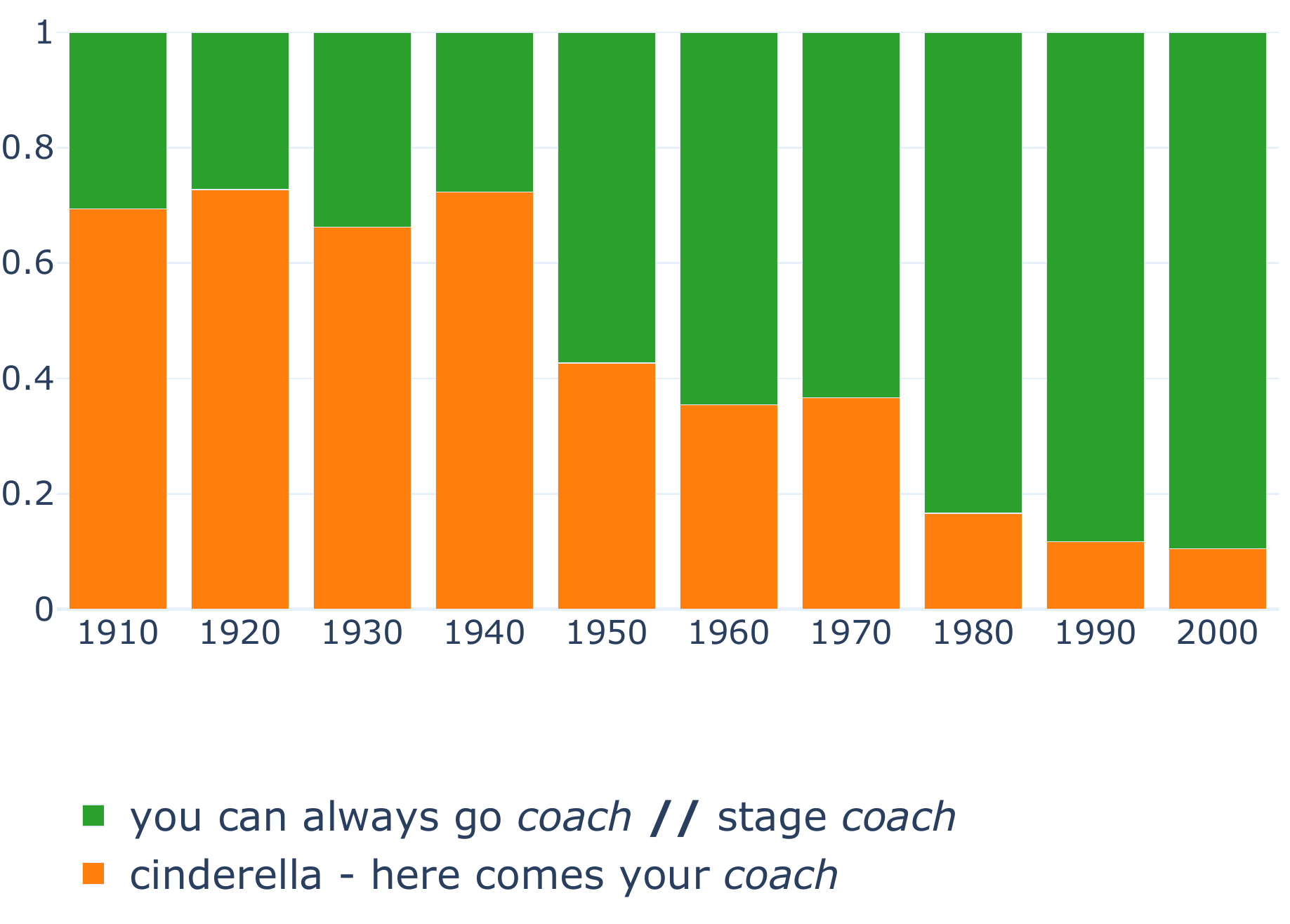}
    \subcaption{\emph{coach}} \label{fig:coach}
 \end{minipage}
 \begin{minipage}{.49\textwidth}
   \centering
   \includegraphics[width=0.95\linewidth]{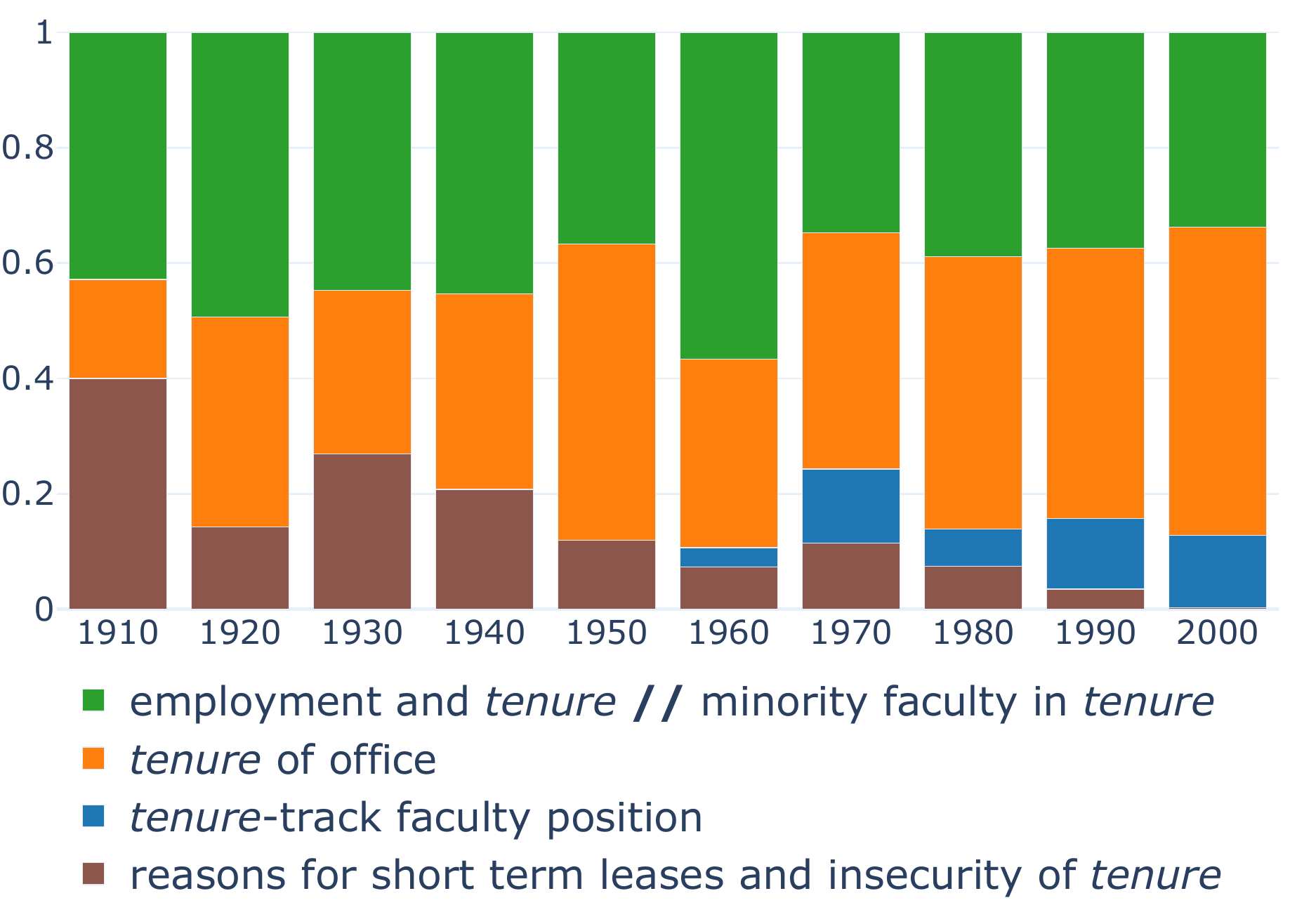}
   \subcaption{\emph{tenure}} \label{fig:tenure}
 \end{minipage} \\ \vspace{1em}
 \begin{minipage}{.49\textwidth}
   \centering
   \includegraphics[width=0.95\linewidth]{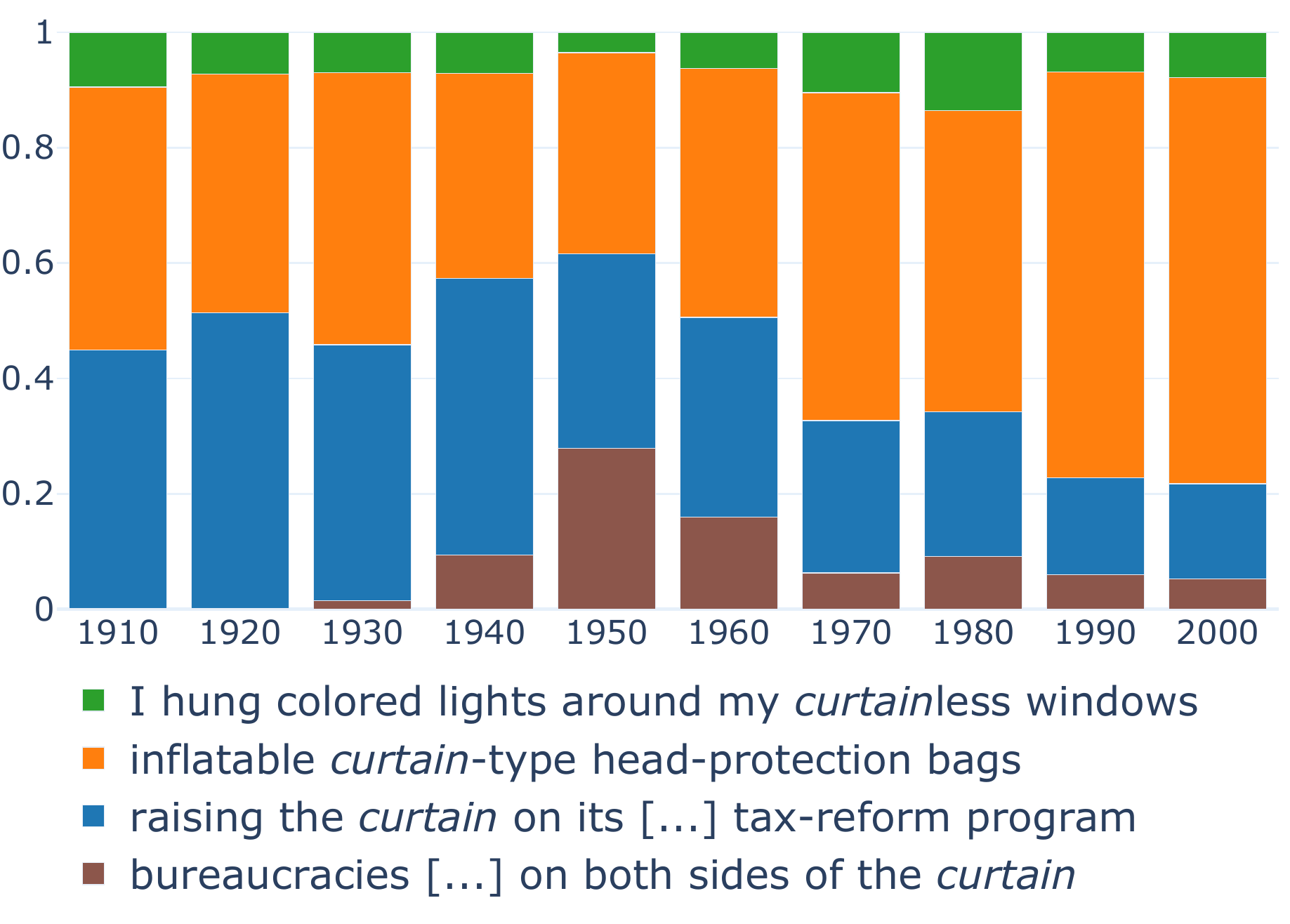}
   \subcaption{\emph{curtain}} \label{fig:curtain} 
 \end{minipage}
 \begin{minipage}{.49\textwidth}
   \centering
   \includegraphics[width=0.95\linewidth]{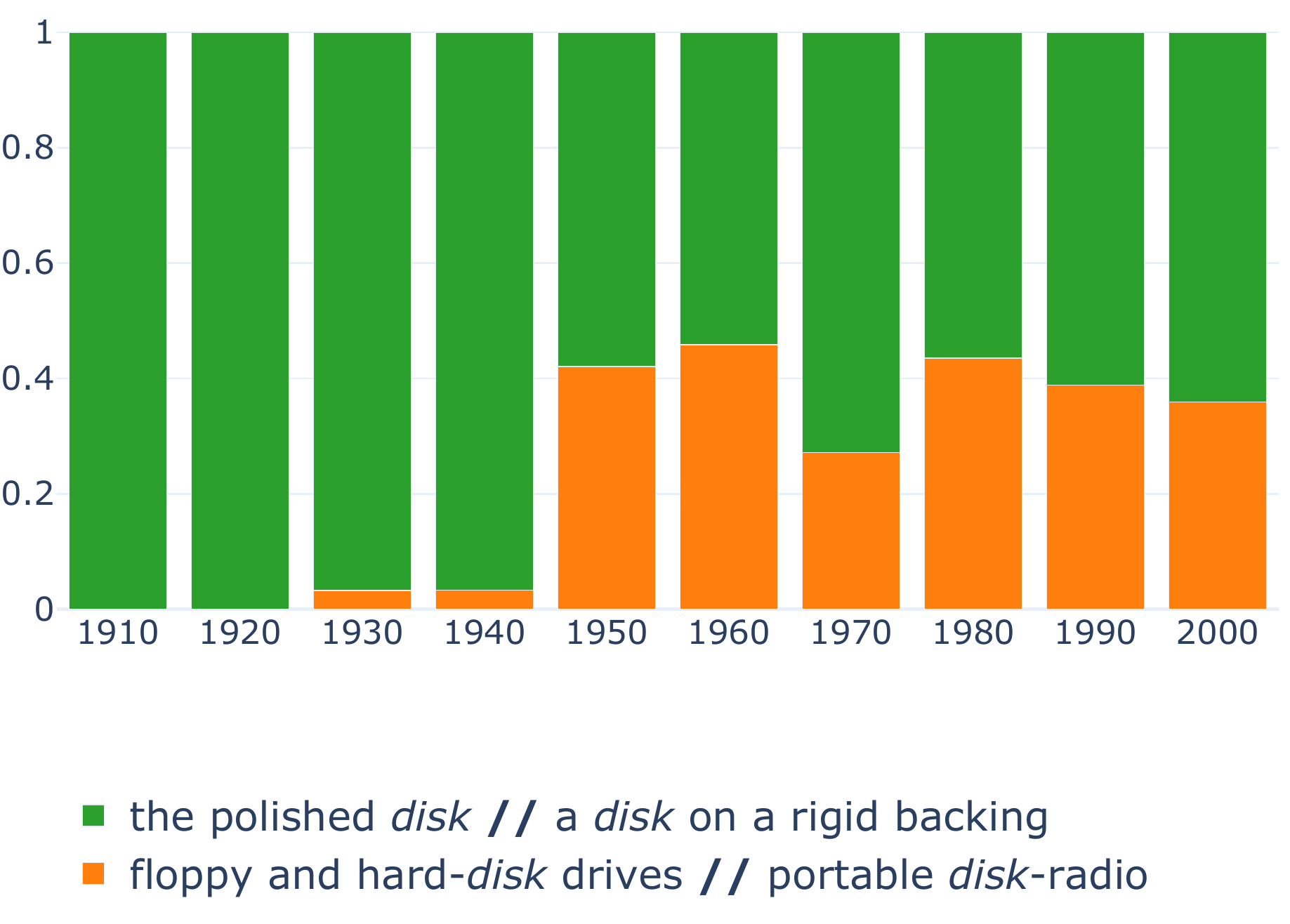}
   \subcaption{\emph{disk}} \label{fig:disk} 
 \end{minipage}
 \caption{Evolution of usage type distributions in the period 1910--2009, generated with occurrences of \textit{coach}, \textit{tenure}, \textit{curtain} and \textit{disk} in COHA \protect\citep{davies2012expanding}. The legends show sample usages per identified usage type.}
 \end{figure*}

\paragraph{Polysemy and homonymy}
Distinctions often occur between underlying senses of polysemous and homonymous words.
For example, the vectors collected for the polysemous word \textit{curious} are grouped together into two usage types, depending on whether \textit{curious} is used to describe something that excites attention as odd, novel, or unexpected (`a wonderful and \textit{curious} and unbelievable story') or rather to describe someone who is marked by a desire to investigate and learn (`\textit{curious} and amazed and innocent'). The same happens for the homonymous usages of the word \textit{coach}, for instance, which can denote vehicles as well as instructors (see Figure \ref{fig:coach} for a diachronic view of the usage types).

\paragraph{Metaphor and metonymy}
In several cases, literal and metaphorical usages are also separated.  
For example, occurrences of \textit{curtain} are clustered into four usage types (Figure \ref{fig:curtain}): two of these correspond to a literal interpretation of the word as a hanging piece of cloth (`\textit{curtain}less windows', `pulled the \textit{curtain} closed') whereas the other two indicate metaphorical interpretations of \textit{curtain} as any barrier that excludes the free exchange of information or communication (`the \textit{curtain} on the legal war is being raised'). 
Similarly, we obtain two usage types for \textit{sphere}: one for literal usages that denote a round solid figure (`the \textit{sphere} of the moon'), and the other for metaphorical interpretations of the word as an area of knowledge or activity (`a certain \textit{sphere} of autonomy') as well as metonymical usages that refer to the planet Earth (`land and peoples on the top half of the \textit{sphere}'). 

\paragraph{Syntactic roles and argument structure}
Further distinctions are observed between word usages that fulfil a different syntactic functionality: not only is part-of-speech ambiguity detected (e.g., `the \textit{cost}-tapered average tariff' vs.\ `\textit{cost} less to make') but contextualised representations also capture regularities in syntactic argument structures. For example, usages of \textit{refuse} are clustered into nominal usages (`society's emotional \textit{refuse}', `the amount of \textit{refuse}'), verbal transitive and intransitive usages (`fall, give up, \textit{refuse}, kick'), as well as verbal usages with infinitive complementation (`\textit{refuse} to go', `\textit{refuse} for the present to sign a treaty').

\paragraph{Collocations and named entities}
Specific clusters are also assigned to lexical items that are parts of phrasal collocations (e.g., `iron \textit{curtain}') or of named entities (`alexander graham \textit{bell}' vs.\ `\textit{bell}-like whistle'). 

\paragraph{Other distinctions}
Some distinctions are interpretable but unexpected. As an example, the word \textit{doubt} does not show the default noun-verb separation but rather a distinction between usages in affirmative contexts (`there is still \textit{doubt}', `the benefit of the \textit{doubt}') and in negative contexts (`there is not a bit of \textit{doubt}', `beyond a reasonable \textit{doubt}'). 

\paragraph{Observed errors}
For some words, we find that usages which appear to be identical are separated into different usage types. In a handful of cases, this seems due to the setup we have used for experimentation, which sets the minimum number of clusters to 2 (see Section~\ref{sec:method:clustering}). This leads to distinct usage types for words such as \textit{maybe}, for which a single type is expected. 
In other cases, a given interpretation is not identified as an independent type, and its usages appear in different clusters. This holds, for example, for the word \textit{tenure}, whose usages in phrases such as `\textit{tenure}-track faculty position' are present in two distinct usage types (see Figure~\ref{fig:tenure}).

Finally, we see that in some cases a usage type ends up including two interpretations which arguably should have been distinguished. For example, two of the usage types identified for \textit{address} are interpretable and coherent: one includes usages in the sense of formal speech and the other one includes verbal usages. The third usage type, however, includes a mix of nominal usages of the word as in `disrespectful manners or \textit{address}' as well as in `network \textit{address}'.

\subsection{What Kinds of Change are Observed?}
\label{sec:analysis:diachronic}
Here we consider usage types diachronically. Different kinds of change, driven by cultural and technological innovation as well as by historical events, emerge from a qualitative inspection of usage distributions along the temporal dimension. 
We describe the most prominent kinds---narrowing and broadening, including metaphorisation---and discuss the extent to which our metrics are able to detect them. 

\paragraph{Narrowing}
Examination of the dynamics of usage distributions allows us to see that for a few words certain usage types disappear or become less common over time (i.e., the interpretation of the word becomes `narrower', less varied). This is the case, for example, for  \textit{coach}, where the frequency decrease of a usage type is gradual and caused by technological evolution (see Figure~\ref{fig:coach}).

Negative mean ED (see Section \ref{sec:method:metrics}) reliably indicates this kind of narrowing. Indeed \textit{coach} is assigned one of the lowest ED score among the GEMS words. 
In contrast, ED fails to detect the obsolescence of a usage type when new usage types emerge simultaneously (since this may lead to no entropy reduction). This is the case, e.g., of \textit{tenure}. The usage type capturing \textit{tenure} of a landed property becomes obsolete; however, we obtain a positive mean ED caused by the appearance of a new usage type (the third type in Figure~\ref{fig:tenure}).

\paragraph{Broadening}
For a substantial amount of words, we observe the emergence of new usage types (i.e., a `broadening' of their use).  This may be due to technological advances as well as to specific historical events. 
As an example, Figure \ref{fig:disk} shows how, starting from the 1950's and as a result of technological innovation, the word \textit{disk} starts to be used to denote also optical disks while beforehand it referred only to generic flat circular objects. 

A special kind of broadening is metaphorisation. As mentioned in Section~\ref{sec:analysis:synchronic}, 
the usage types for the word \textit{curtain} include metaphorical interpretations. Figure \ref{fig:curtain} allows us to see when the metaphorical meaning related to the historically charged expression \textit{iron curtain} is acquired. This novel usage type is related to a specific historical period: it emerges between the 1930's and the 1940's, reaches its peak in the 1950's, and remains stably low in frequency starting from the 1970's.

The metrics that best capture broadening are JSD and APD---e.g., \textit{disk} is assigned a high semantic change score by both metrics. Yet, sometimes these metrics generate different score rankings. For example, \textit{curtain} yields a rather low APD score due to the low relative frequency of the novel usage (Figure \ref{fig:curtain}). In contrast, even though the novel usage type is not very prominent in some decades, JSD can still discriminate it and measure its development.
On the other hand, the word \textit{address}, for which we also observe broadening, is assigned a low score by JSD due to the errors in its usage type assignments pointed out in Section~\ref{sec:analysis:synchronic}. As APD does not rely on usage types, it is not affected by this issue and does indeed assign a high change score to the word. 

Finally, although our metrics help us identify the broadening of a word's meaning, they cannot capture the type of broadening (i.e., the nature of the emerging interpretations). Detecting metaphorisation, for example, may require inter-cluster comparisons to identify a metaphor's source and target usage types, which we leave to future work.


\section{Conclusion}
\label{sec:conclusion}
We have introduced a novel approach to the analysis of lexical semantic change. To our knowledge, this is the first work that tackles this problem using neural contextualised word representations and no lexicographic supervision.
We have shown that the representations and the detected semantic shifts are aligned to human interpretation, and presented a new dataset of human similarity judgements which can be used to measure said alignment. Finally, through extensive qualitative analysis, we have demonstrated that our method allows us to capture a variety of synchronic and diachronic linguistic phenomena.

Our approach offers several advantages over previous methods: (1) it does not rely on a fixed number of word senses, (2) it captures morphosyntactic properties of word usage, and (3) it offers a more effective interpretation of lexical meaning by enabling the inspection of particular example sentences.
In recent work, we have experimented with alternative ways of obtaining usage representations (using a different language model, fine-tuning, and various layer selection strategies) and we have obtained very promising results in detecting semantic change across four languages \citep{kutuzov2020}.
In the future, we plan to investigate whether usage representations can provide an even finer grained account of lexical meaning and its dynamics, e.g., to automatically discriminate between different types of meaning change. We expect our work to inspire further analyses of variation and change which exploit the expressiveness of contextualised word representations.

\section*{Acknowledgments}
This paper builds upon the preliminary work presented by \citet{giulianelli2019thesis}. We would like to thank Lisa Beinborn for providing useful feedback as well as the three anonymous ACL reviewers for their helpful comments. This project has received funding from the European Research Council (ERC) under the European Union’s Horizon 2020 research and innovation programme (grant agreement No.~819455).

\bibliography{msc-thesis}

\begin{thebibliography}{58}
\expandafter\ifx\csname natexlab\endcsname\relax\def\natexlab#1{#1}\fi

\bibitem[{Arthur and Vassilvitskii(2007)}]{arthur2007k}
David Arthur and Sergei Vassilvitskii. 2007.
\newblock \texttt{k-means++}: {T}he {A}dvantages of {C}areful {S}eeding.
\newblock In \emph{Proceedings of the Eighteenth Annual ACM-SIAM Symposium on
  Discrete Algorithms}, pages 1027--1035. Society for Industrial and Applied
  Mathematics.

\bibitem[{Bamler and Mandt(2017)}]{bamler2017dynamic}
Robert Bamler and Stephan Mandt. 2017.
\newblock {D}ynamic {W}ord {E}mbeddings.
\newblock In \emph{Proceedings of the 34th International Conference on Machine
  Learning-Volume 70}, pages 380--389. JMLR.org.

\bibitem[{Brown(2008)}]{brown-2008-choosing}
Susan~Windisch Brown. 2008.
\newblock {C}hoosing {S}ense {D}istinctions for {WSD}: {P}sycholinguistic
  {E}vidence.
\newblock In \emph{Proceedings of ACL-08: HLT, Short Papers}, pages 249--252,
  Columbus, Ohio. Association for Computational Linguistics.

\bibitem[{Brugman(1988)}]{brugman1988lexicon}
Claudia~Marlea Brugman. 1988.
\newblock \emph{{T}he {S}tory of {O}ver: {P}olysemy, {S}emantics, and the
  {S}tructure of the {L}exicon}.
\newblock Garland, New York.

\bibitem[{Bybee(2015)}]{bybee2015language}
Joan Bybee. 2015.
\newblock \emph{{L}anguage {C}hange}.
\newblock Cambridge University Press.

\bibitem[{Cook et~al.(2014)Cook, Lau, McCarthy, and Baldwin}]{cook2014}
Paul Cook, Jey~Han Lau, Diana McCarthy, and Timothy Baldwin. 2014.
\newblock {N}ovel {W}ord-{S}ense {I}dentification.
\newblock In \emph{Proceedings of COLING 2014, the 25th International
  Conference on Computational Linguistics: Technical Papers}, pages 1624--1635.

\bibitem[{Dai and Le(2015)}]{dai2015semi}
Andrew~M Dai and Quoc~V Le. 2015.
\newblock {S}emi-supervised {S}equence {L}earning.
\newblock In \emph{Advances in Neural Information Processing Systems}, pages
  3079--3087.

\bibitem[{Davies(2012)}]{davies2012expanding}
Mark Davies. 2012.
\newblock {E}xpanding {H}orizons in {H}istorical {L}inguistics with the
  400-{M}illion {W}ord {C}orpus of {H}istorical {A}merican {E}nglish.
\newblock \emph{Corpora}, 7(2):121--157.

\bibitem[{Del~Tredici et~al.(2019)Del~Tredici, Fern{\'a}ndez, and
  Boleda}]{deltredici2019short}
Marco Del~Tredici, Raquel Fern{\'a}ndez, and Gemma Boleda. 2019.
\newblock {S}hort-{T}erm {M}eaning {S}hift: {A} {D}istributional {E}xploration.
\newblock In \emph{Proceedings of NAACL-HLT 2019 (Annual Conference of the
  North American Chapter of the Association for Computational Linguistics)}.

\bibitem[{Devlin et~al.(2019)Devlin, Chang, Lee, and Toutanova}]{devlin2018}
Jacob Devlin, Ming-Wei Chang, Kenton Lee, and Kristina Toutanova. 2019.
\newblock {BERT}: {P}re-training of {D}eep {B}idirectional {T}ransformers for
  {L}anguage {U}nderstanding.
\newblock In \emph{Proceedings of the 2019 Conference of the North American
  Chapter of the Association for Computational Linguistics: Human Language
  Technologies, Volume 1 (Long and Short Papers)}, pages 4171--4186.

\bibitem[{Dubossarsky et~al.(2015)Dubossarsky, Tsvetkov, Dyer, and
  Grossman}]{dubossarsky2015category}
Haim Dubossarsky, Yulia Tsvetkov, Chris Dyer, and Eitan Grossman. 2015.
\newblock {A} {B}ottom {U}p {A}pproach to {C}ategory {M}apping and {M}eaning
  {C}hange.
\newblock In \emph{Word Structure and Word Usage. Proceedings of the NetWordS
  Final Conference}, pages 66--70.

\bibitem[{Erk et~al.(2009)Erk, McCarthy, and Gaylord}]{erk2009investigations}
Katrin Erk, Diana McCarthy, and Nicholas Gaylord. 2009.
\newblock {I}nvestigations on {W}ord {S}enses and {W}ord {U}sages.
\newblock In \emph{Proceedings of the Joint Conference of the 47th Annual
  Meeting of the {ACL} and the 4th International Joint Conference on Natural
  Language Processing of the {AFNLP}}, pages 10--18, Suntec, Singapore.
  Association for Computational Linguistics.

\bibitem[{Erk et~al.(2013)Erk, McCarthy, and Gaylord}]{erk2013measuring}
Katrin Erk, Diana McCarthy, and Nicholas Gaylord. 2013.
\newblock {M}easuring {W}ord {M}eaning in {C}ontext.
\newblock \emph{Computational Linguistics}, 39(3):511--554.

\bibitem[{Erk and Pad{\'o}(2008)}]{erk2008structured}
Katrin Erk and Sebastian Pad{\'o}. 2008.
\newblock {A} {S}tructured {V}ector {S}pace {M}odel for {W}ord {M}eaning in
  {C}ontext.
\newblock In \emph{Proceedings of the 2008 Conference on Empirical Methods in
  Natural Language Processing}, pages 897--906.

\bibitem[{Erk and Pad{\'o}(2010)}]{erk2010exemplar}
Katrin Erk and Sebastian Pad{\'o}. 2010.
\newblock {E}xemplar-{B}ased {M}odels for {W}ord {M}eaning in {C}ontext.
\newblock In \emph{Proceedings of the ACL 2010 Conference (Short Papers)},
  pages 92--97.

\bibitem[{Frermann and Lapata(2016)}]{frermann2016}
Lea Frermann and Mirella Lapata. 2016.
\newblock {A} {B}ayesian {M}odel of {D}iachronic {M}eaning {C}hange.
\newblock \emph{Transactions of the Association for Computational Linguistics},
  4:31--45.

\bibitem[{Giulianelli(2019)}]{giulianelli2019thesis}
Mario Giulianelli. 2019.
\newblock {L}exical {S}emantic {C}hange {A}nalysis with {C}ontextualised {W}ord
  {R}epresentations.
\newblock Master's thesis, University of Amsterdam, July.

\bibitem[{Gulordava and Baroni(2011)}]{gulordava2011}
Kristina Gulordava and Marco Baroni. 2011.
\newblock {A} {D}istributional {S}imilarity {A}pproach to the {D}etection of
  {S}emantic {C}hange in the {G}oogle {B}ooks {N}gram {C}orpus.
\newblock In \emph{Proceedings of the GEMS 2011 Workshop on Geometrical Models
  of Natural Language Semantics}, pages 67--71.

\bibitem[{Hamilton et~al.(2016)Hamilton, Leskovec, and
  Jurafsky}]{hamilton2016diachronic}
William~L Hamilton, Jure Leskovec, and Dan Jurafsky. 2016.
\newblock {D}iachronic {W}ord {E}mbeddings {R}eveal {S}tatistical {L}aws of
  {S}emantic {C}hange.
\newblock In \emph{Proceedings of the 54th Annual Meeting of the Association
  for Computational Linguistics (Volume 1: Long Papers)}, pages 1489--1501.

\bibitem[{Hopper et~al.(1991)}]{hopper1991grammaticization}
Paul~J Hopper et~al. 1991.
\newblock {O}n {S}ome {P}rinciples of {G}rammaticization.
\newblock \emph{Approaches to Grammaticalization}, 1:17--35.

\bibitem[{Howard and Ruder(2018)}]{howard2018universal}
Jeremy Howard and Sebastian Ruder. 2018.
\newblock {U}niversal {L}anguage {M}odel {F}ine-tuning for {T}ext
  {C}lassification.
\newblock In \emph{Proceedings of the 56th Annual Meeting of the Association
  for Computational Linguistics (Volume 1: Long Papers)}, pages 328--339.

\bibitem[{Hu et~al.(2019)Hu, Li, and Liang}]{hu2019diachronic}
Renfen Hu, Shen Li, and Shichen Liang. 2019.
\newblock {D}iachronic {S}ense {M}odeling with {D}eep {C}ontextualized {W}ord
  {E}mbeddings: {A}n {E}cological {V}iew.
\newblock In \emph{Proceedings of the 57th Annual Meeting of the Association
  for Computational Linguistics}, pages 3899--3908, Florence, Italy.
  Association for Computational Linguistics.

\bibitem[{Kilgarriff(1997)}]{kilgarriff1997}
Adam Kilgarriff. 1997.
\newblock {I} {D}on't {B}elieve in {W}ord {S}enses.
\newblock \emph{Computers and the Humanities}, 31(2):91--113.

\bibitem[{Kim et~al.(2014)Kim, Chiu, Hanaki, Hegde, and Petrov}]{kim2014}
Yoon Kim, Yi-I Chiu, Kentaro Hanaki, Darshan Hegde, and Slav Petrov. 2014.
\newblock {T}emporal {A}nalysis of {L}anguage through {N}eural {L}anguage
  {M}odels.
\newblock In \emph{Proceedings of the ACL 2014 Workshop on Language
  Technologies and Computational Social Science}, pages 61--65.

\bibitem[{Kulkarni et~al.(2015)Kulkarni, Al-Rfou, Perozzi, and
  Skiena}]{kulkarni2015}
Vivek Kulkarni, Rami Al-Rfou, Bryan Perozzi, and Steven Skiena. 2015.
\newblock {S}tatistically {S}ignificant {D}etection of {L}inguistic {C}hange.
\newblock In \emph{Proceedings of the 24th International Conference on World
  Wide Web}, pages 625--635. International World Wide Web Conferences Steering
  Committee.

\bibitem[{Kutuzov and Giulianelli(2020)}]{kutuzov2020}
Andrey Kutuzov and Mario Giulianelli. 2020.
\newblock {U}i{O}-{U}v{A} at {S}em{E}val-2020 {T}ask 1: {C}ontextualised
  {E}mbeddings for {L}exical {S}emantic {C}hange {D}etection.
\newblock Forthcoming.

\bibitem[{Kutuzov et~al.(2018)Kutuzov, {\O}vrelid, Szymanski, and
  Velldal}]{kutuzov2018survey}
Andrey Kutuzov, Lilja {\O}vrelid, Terrence Szymanski, and Erik Velldal. 2018.
\newblock {D}iachronic {W}ord {E}mbeddings and {S}emantic {S}hifts: {A}
  {S}urvey.
\newblock In \emph{Proceedings of the 27th International Conference on
  Computational Linguistics}, pages 1384--1397.

\bibitem[{Lau et~al.(2014)Lau, Cook, McCarthy, Gella, and
  Baldwin}]{lau2014learning}
Jey~Han Lau, Paul Cook, Diana McCarthy, Spandana Gella, and Timothy Baldwin.
  2014.
\newblock {L}earning {W}ord {S}ense {D}istributions, {D}etecting {U}nattested
  {S}enses and {I}dentifying {N}ovel {S}enses {U}sing {T}opic {M}odels.
\newblock In \emph{Proceedings of the 52nd Annual Meeting of the Association
  for Computational Linguistics (Volume 1: Long Papers)}, volume~1, pages
  259--270.

\bibitem[{Lau et~al.(2012)Lau, Cook, McCarthy, Newman, and Baldwin}]{lau2012}
Jey~Han Lau, Paul Cook, Diana McCarthy, David Newman, and Timothy Baldwin.
  2012.
\newblock {W}ord {S}ense {I}nduction for {N}ovel {S}ense {D}etection.
\newblock In \emph{Proceedings of the 13th Conference of the European Chapter
  of the Association for Computational Linguistics}, pages 591--601.
  Association for Computational Linguistics.

\bibitem[{Lin(1991)}]{lin1991jsd}
Jianhua Lin. 1991.
\newblock {D}ivergence {M}easures {B}ased on the {S}hannon {E}ntropy.
\newblock \emph{IEEE Transactions on Information theory}, 37(1):145--151.

\bibitem[{Ludlow(2014)}]{ludlow2014}
Peter Ludlow. 2014.
\newblock \emph{{L}iving {W}ords: {M}eaning {U}nderdetermination and the
  {D}ynamic {L}exicon}.
\newblock OUP Oxford.

\bibitem[{Manandhar et~al.(2010)Manandhar, Klapaftis, Dligach, and
  Pradhan}]{manandhar2010semeval}
Suresh Manandhar, Ioannis~P Klapaftis, Dmitriy Dligach, and Sameer~S Pradhan.
  2010.
\newblock {S}em{E}val-2010 {T}ask 14: {W}ord {S}ense {I}nduction \&
  {D}isambiguation.
\newblock In \emph{Proceedings of the 5th International Workshop on Semantic
  Evaluation}, pages 63--68. Association for Computational Linguistics.

\bibitem[{Mantel(1967)}]{mantel1967}
Nathan Mantel. 1967.
\newblock {T}he {D}etection of {D}isease {C}lustering and a {G}eneralized
  {R}egression {A}pproach.
\newblock \emph{Cancer Research}, 27(2):209--220.

\bibitem[{Martinc et~al.(2020)Martinc, Montariol, Zosa, and
  Pivovarova}]{martinc2020}
Matej Martinc, Syrielle Montariol, Elaine Zosa, and Lidia Pivovarova. 2020.
\newblock {C}apturing {E}volution in {W}ord {U}sage: {J}ust {A}dd {M}ore
  {C}lusters?
\newblock In \emph{Companion Proceedings of the International World Wide Web
  Conference}, pages 20--24.

\bibitem[{McCann et~al.(2017)McCann, Bradbury, Xiong, and
  Socher}]{mccann2017learned}
Bryan McCann, James Bradbury, Caiming Xiong, and Richard Socher. 2017.
\newblock {L}earned in {T}ranslation: {C}ontextualized {W}ord {V}ectors.
\newblock In \emph{Advances in Neural Information Processing Systems}, pages
  6294--6305.

\bibitem[{Michel et~al.(2011)Michel, Shen, Aiden, Veres, Gray, Pickett,
  Hoiberg, Clancy, Norvig, Orwant et~al.}]{michel2011quantitative}
Jean-Baptiste Michel, Yuan~Kui Shen, Aviva~Presser Aiden, Adrian Veres,
  Matthew~K Gray, Joseph~P Pickett, Dale Hoiberg, Dan Clancy, Peter Norvig, Jon
  Orwant, et~al. 2011.
\newblock {Q}uantitative {A}nalysis of {C}ulture {U}sing {M}illions of
  {D}igitized {B}ooks.
\newblock \emph{Science}, 331(6014):176--182.

\bibitem[{Mitra et~al.(2015)Mitra, Mitra, Maity, Riedl, Biemann, Goyal, and
  Mukherjee}]{mitra2015}
Sunny Mitra, Ritwik Mitra, Suman~Kalyan Maity, Martin Riedl, Chris Biemann,
  Pawan Goyal, and Animesh Mukherjee. 2015.
\newblock {A}n {A}utomatic {A}pproach to {I}dentify {W}ord {S}ense {C}hanges in
  {T}ext {M}edia across {T}imescales.
\newblock \emph{Natural Language Engineering}, 21(5):773--798.

\bibitem[{Mitra et~al.(2014)Mitra, Mitra, Riedl, Biemann, Mukherjee, and
  Goyal}]{mitra2014}
Sunny Mitra, Ritwik Mitra, Martin Riedl, Chris Biemann, Animesh Mukherjee, and
  Pawan Goyal. 2014.
\newblock {T}hat's {S}ick {D}ude! {A}utomatic {I}dentification of {W}ord
  {S}ense {C}hange across {D}ifferent {T}imescales.
\newblock In \emph{Proceedings of the 52nd Annual Meeting of the Association
  for Computational Linguistics (Volume 1: Long Papers)}, pages 1020--1029.

\bibitem[{Navigli and Vannella(2013)}]{navigli2013semeval}
Roberto Navigli and Daniele Vannella. 2013.
\newblock {S}em{E}val-2013 {T}ask 11: {W}ord {S}ense {I}nduction and
  {D}isambiguation within an {E}nd-{U}ser {A}pplication.
\newblock In \emph{Second Joint Conference on Lexical and Computational
  {S}emantics (* SEM), Volume 2: Proceedings of the Seventh International
  Workshop on Semantic Evaluation (SemEval 2013)}, pages 193--201.

\bibitem[{Pantel and Lin(2002)}]{pantel2002}
Patrick Pantel and Dekang Lin. 2002.
\newblock {D}iscovering {W}ord {S}enses from {T}ext.
\newblock In \emph{Proceedings of the Eighth ACM SIGKDD International
  Conference on Knowledge Discovery and Data Mining}, KDD ’02, page
  613–619, New York, NY, USA. Association for Computing Machinery.

\bibitem[{Paradis(2011)}]{paradis2011metonymization}
Carita Paradis. 2011.
\newblock {M}etonymization: {A} {K}ey {M}echanism in {S}emantic {C}hange.
\newblock \emph{Defining Metonymy in Cognitive Linguistics: Towards a Consensus
  View}, pages 61--98.

\bibitem[{Peters et~al.(2018)Peters, Neumann, Iyyer, Gardner, Clark, Lee, and
  Zettlemoyer}]{peters2018}
Matthew Peters, Mark Neumann, Mohit Iyyer, Matt Gardner, Christopher Clark,
  Kenton Lee, and Luke Zettlemoyer. 2018.
\newblock {D}eep {C}ontextualized {W}ord {R}epresentations.
\newblock In \emph{Proceedings of the 2018 Conference of the North American
  Chapter of the Association for Computational Linguistics: Human Language
  Technologies, Volume 1 (Long Papers)}, pages 2227--2237.

\bibitem[{Pilehvar and Camacho-Collados(2019)}]{wic2018}
Mohammad~Taher Pilehvar and Jose Camacho-Collados. 2019.
\newblock {W}i{C}: the {W}ord-in-{C}ontext {D}ataset for {E}valuating
  {C}ontext-{S}ensitive {M}eaning {R}epresentations.
\newblock In \emph{Proceedings of the 2019 Conference of the North {A}merican
  Chapter of the Association for Computational Linguistics: Human Language
  Technologies, Volume 1 (Long and Short Papers)}, pages 1267--1273,
  Minneapolis, Minnesota. Association for Computational Linguistics.

\bibitem[{Radford et~al.(2018)Radford, Narasimhan, Salimans, and
  Sutskever}]{radford2018}
Alec Radford, Karthik Narasimhan, Tim Salimans, and Ilya Sutskever. 2018.
\newblock {I}mproving {L}anguage {U}nderstanding by {G}enerative
  {P}re-training.
\newblock Technical report, OpenAI.

\bibitem[{Radford et~al.(2019)Radford, Wu, Child, Luan, Amodei, and
  Sutskever}]{radford2019language}
Alec Radford, Jeffrey Wu, Rewon Child, David Luan, Dario Amodei, and Ilya
  Sutskever. 2019.
\newblock {L}anguage {M}odels are {U}nsupervised {M}ultitask {L}earners.
\newblock Technical report, OpenAI.

\bibitem[{R{\'e} and Azad(2014)}]{re2014generalization}
Miguel~A R{\'e} and Rajeev~K Azad. 2014.
\newblock {G}eneralization of {E}ntropy {B}ased {D}ivergence {M}easures for
  {S}ymbolic {S}equence {A}nalysis.
\newblock \emph{PloS one}, 9(4):e93532.

\bibitem[{Rosenfeld and Erk(2018)}]{rosenfeld2018deep}
Alex Rosenfeld and Katrin Erk. 2018.
\newblock {D}eep {N}eural {M}odels of {S}emantic {S}hift.
\newblock In \emph{Proceedings of the 2018 Conference of the North American
  Chapter of the Association for Computational Linguistics: Human Language
  Technologies, Volume 1 (Long Papers)}, pages 474--484.

\bibitem[{Rousseeuw(1987)}]{rousseeuw1987silhouettes}
Peter~J. Rousseeuw. 1987.
\newblock {S}ilhouettes: {A} {G}raphical {A}id to the {I}nterpretation and
  {V}alidation of {C}luster {A}nalysis.
\newblock \emph{Journal of Computational and Applied Mathematics}, 20:53--65.

\bibitem[{Rudolph and Blei(2018)}]{rudolph2018}
Maja Rudolph and David Blei. 2018.
\newblock Dynamic embeddings for {L}anguage evolution.
\newblock In \emph{Proceedings of the 2018 World Wide Web Conference on World
  Wide Web}, pages 1003--1011. International World Wide Web Conferences
  Steering Committee.

\bibitem[{Schlechtweg et~al.(2018)Schlechtweg, Schulte~im Walde, and
  Eckmann}]{schlechtweg2018durel}
Dominik Schlechtweg, Sabine Schulte~im Walde, and Stefanie Eckmann. 2018.
\newblock {D}iachronic {U}sage {R}elatedness ({DUR}el): {A} {F}ramework for the
  {A}nnotation of {L}exical {S}emantic {C}hange.
\newblock In \emph{Proceedings of the 2018 Conference of the North American
  Chapter of the Association for Computational Linguistics: Human Language
  Technologies, Volume 2 (Short Papers)}, volume~2, pages 169--174.

\bibitem[{Sch{\"u}tze(1998)}]{schutze1998automatic}
Hinrich Sch{\"u}tze. 1998.
\newblock {A}utomatic {W}ord {S}ense {D}iscrimination.
\newblock \emph{Computational Linguistics}, 24(1):97--123.

\bibitem[{Tang(2018)}]{tang2018survey}
Xuri Tang. 2018.
\newblock {A} {S}tate-of-the-{A}rt of {S}emantic {C}hange {c}omputation.
\newblock \emph{Natural Language Engineering}, 24(5):649--676.

\bibitem[{Tang et~al.(2016)Tang, Qu, and Chen}]{tang2016semantic}
Xuri Tang, Weiguang Qu, and Xiaohe Chen. 2016.
\newblock {S}emantic {C}hange {C}omputation: {A} {S}uccessive {A}pproach.
\newblock \emph{World Wide Web}, 19(3):375--415.

\bibitem[{Vaswani et~al.(2017)Vaswani, Shazeer, Parmar, Uszkoreit, Jones,
  Gomez, Kaiser, and Polosukhin}]{vaswani2017attention}
Ashish Vaswani, Noam Shazeer, Niki Parmar, Jakob Uszkoreit, Llion Jones,
  Aidan~N Gomez, {\L}ukasz Kaiser, and Illia Polosukhin. 2017.
\newblock {A}ttention {I}s {A}ll {Y}ou {N}eed.
\newblock In \emph{Advances in Neural Information Processing Systems}, pages
  5998--6008.

\bibitem[{Wiedemann et~al.(2019)Wiedemann, Remus, Chawla, and
  Biemann}]{wiedemann2019does}
Gregor Wiedemann, Steffen Remus, Avi Chawla, and Chris Biemann. 2019.
\newblock {D}oes {BERT} {M}ake {A}ny {S}ense? {I}nterpretable {W}ord {S}ense
  {D}isambiguation with {C}ontextualized {E}mbeddings.
\newblock In \emph{Proceedings of the 15th Conference on Natural Language
  Processing, {KONVENS} 2019}, Erlangen, Germany.

\bibitem[{Wijaya and Yeniterzi(2011)}]{wijaya2011understanding}
Derry~Tanti Wijaya and Reyyan Yeniterzi. 2011.
\newblock {U}nderstanding {S}emantic {C}hange of {W}ords over {C}enturies.
\newblock In \emph{Proceedings of the 2011 International Workshop on Detecting
  and Exploiting Cultural Diversity on the Social Web}, pages 35--40. ACM.

\bibitem[{Xu and Kemp(2015)}]{xu2015laws}
Yang Xu and Charles Kemp. 2015.
\newblock {A} {C}omputational {E}valuation of {T}wo {L}aws of {S}emantic
  {C}hange.
\newblock In \emph{CogSci}.

\bibitem[{Zhu et~al.(2015)Zhu, Kiros, Zemel, Salakhutdinov, Urtasun, Torralba,
  and Fidler}]{zhu2015books}
Yukun Zhu, Ryan Kiros, Rich Zemel, Ruslan Salakhutdinov, Raquel Urtasun,
  Antonio Torralba, and Sanja Fidler. 2015.
\newblock {A}ligning {B}ooks and {M}ovies: {T}owards {S}tory-{L}ike {V}isual
  {E}xplanations by {W}atching {M}ovies and {R}eading {B}ooks.
\newblock In \emph{Proceedings of the IEEE International Conference on Computer
  Vision}, pages 19--27.

\end{thebibliography}
\bibliographystyle{acl_natbib}

\appendix

\section{Appendix}

This appendix includes supplementary materials related to Section \ref{sec:eval:similarities}.

\subsection{New Dataset of Similarity Judgements} 
\label{appendix-newdataset}

\paragraph{Obtaining usage pairs}
For each of our 16 target words, we sample five usages from each of their usage types, one for every 20-year period in the last century of COHA. When a usage type does not occur in a time interval, we uniformly sample an interval from those that do contain occurrences of that usage type. All possible pairwise combinations (without replacement) are generated for each target word, resulting in a total of 3,285 usage pairs.

\paragraph{Crowdsourced annotation}
We use the crowdsourcing platform Figure Eight (since then acquired by Appen\footnote{\url{https://appen.com}}) to collect five similarity judgements for each of these usage pairs. To control the quality of the similarity judgements, we select Figure Eight workers from the pool of most experienced contributors, we require them to be native English speakers and to have completed a test quiz consisting of 10 similarity judgements. For this purpose, 170 usage pairs were manually annotated by the first author with 1 to 3 acceptable labels. The compensation scheme for the raters is based on an average wage of 10 USD per hour.

\begin{figure}[h!]
    \centering
    \includegraphics[width=0.5\textwidth]{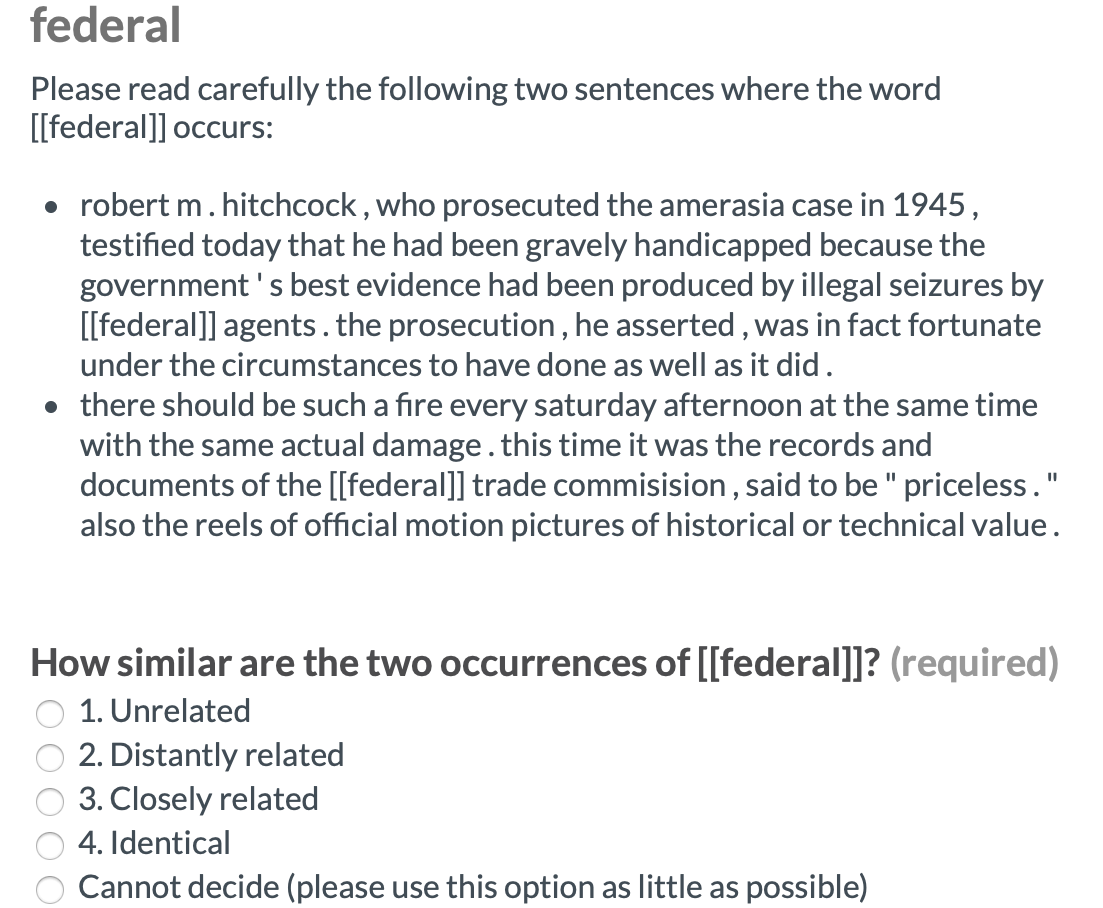}
    \caption{An annotation item on the Figure Eight crowdsourcing platform.}
    \label{fig:annotation}
\end{figure}

\begin{figure*}
    \centering
    \includegraphics[width=\textwidth]{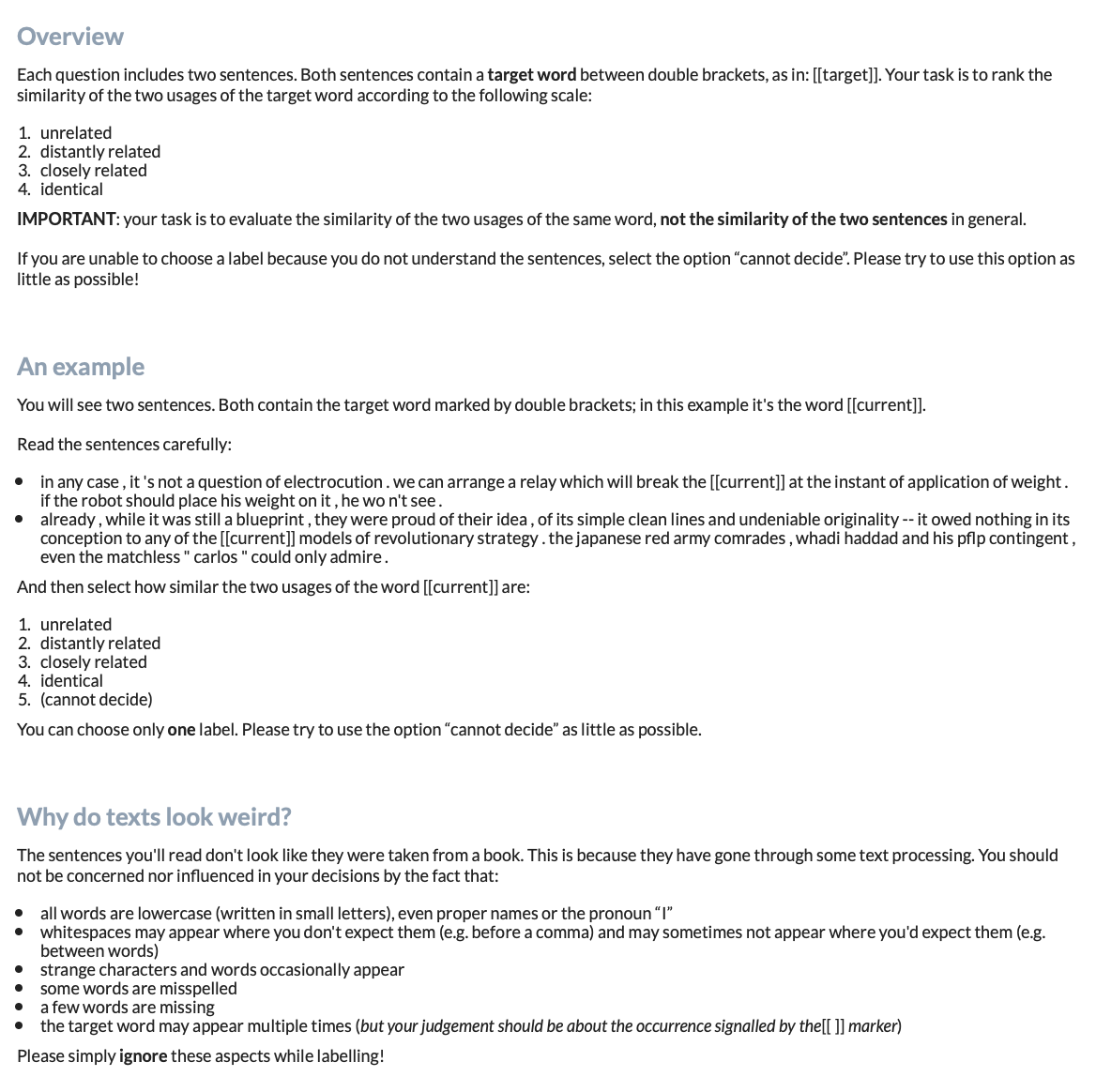}
    \caption{Annotation instructions (part 1).}
    \label{fig:instr1}
\end{figure*}

\begin{figure*}
    \centering
    \includegraphics[width=\textwidth]{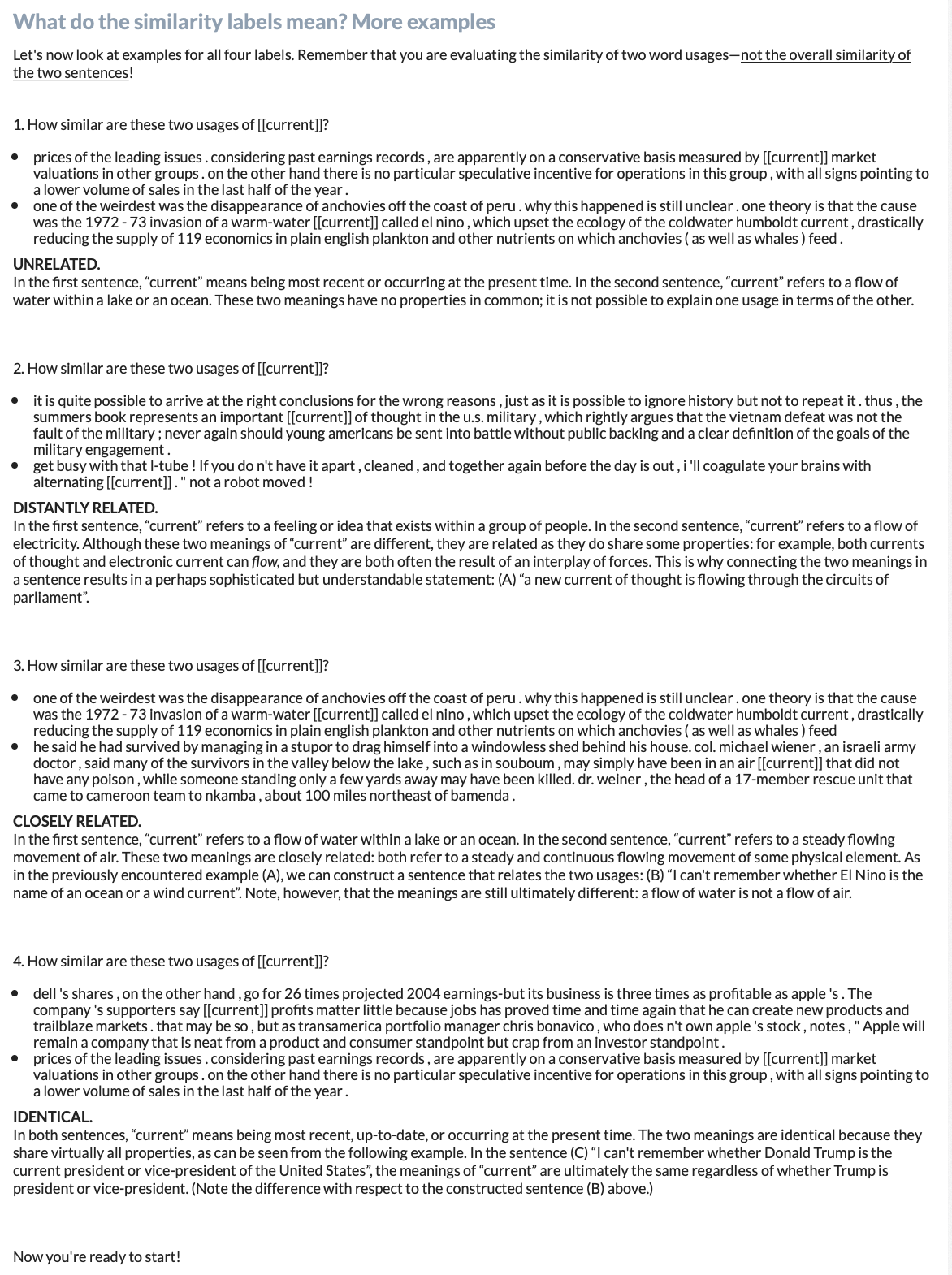}
    \caption{Annotation instructions (part 2).}
    \label{fig:instr2}
\end{figure*}

Figures \ref{fig:instr1} and \ref{fig:instr2} (on the next pages) show the full instructions given to the annotators and Figure \ref{fig:annotation} illustrates a single annotation item.

\subsection{Correlation Results}
\label{appendix-correlation}
We measure Spearman’s rank correlation between human- and machine-generated usage similarity matrices using the Mantel test and observe a significant positive correlation for 10 out of 16 words. Table~\ref{tab:human-bert-correlation} presents the correlation coefficients and $p$-values obtained for each word.

\begin{table}[h!]
\centering
\resizebox{0.5\columnwidth}{!}{
\begin{tabular}{rcc}
\textbf{} & $\rho$ & $p$ \\ \hline
federal    & 0.131 & 0.001 \\
spine      & 0.195 & 0.032 \\
optical    & 0.227 & 0.003 \\
compact    & 0.229 & 0.002 \\
signal     & 0.233 & 0.008 \\
leaf       & 0.252 & 0.001 \\
net        & 0.361 & 0.001 \\
coach      & 0.433 & 0.007 \\
sphere     & 0.446 & 0.002 \\
mirror     & 0.454 & 0.027 \\
\hline
card       & 0.358 & 0.055 \\
virus      & 0.271 & 0.159 \\
disk       & 0.183 & 0.211 \\
brick      & 0.203 & 0.263 \\
virtual    & -0.085 & 0.561 \\
energy     & 0.002 & 0.990 \\       
\end{tabular}
}
\caption{Correlation results per word.}
\label{tab:human-bert-correlation}
\end{table}


\end{document}